\documentclass[10pt,twocolumn,letterpaper]{article}

\usepackage{iccv}
\usepackage{times}
\usepackage{epsfig}
\usepackage{graphicx}
\usepackage{amsmath}
\usepackage{amssymb}
\usepackage{overpic}
\usepackage{bm}
\usepackage{paralist}
\usepackage{multirow}
\usepackage{textcomp}
\usepackage{pifont}
\usepackage[accsupp]{axessibility}  

\def\blu#1{\textbf{\color{blue} #1}} 
\def\red#1{\textbf{\color{red}\underline{#1}}} 

\usepackage[pagebackref=true,breaklinks=true,letterpaper=true,colorlinks,bookmarks=false]{hyperref}

\iccvfinalcopy 

\def\iccvPaperID{8492} 

\ificcvfinal\fi

\begin{document}

\title{Summarize and Search: Learning Consensus-aware Dynamic Convolution for Co-Saliency Detection}

\author{
	Ni Zhang$^{1}$
	\hspace{25pt}
	Junwei Han$^{1}$
	\hspace{25pt}
	Nian Liu$^{2}$\footnotemark[1]
	\hspace{25pt}
	Ling Shao$^{2}$
	\hspace{25pt}
	\\
	$^1$Northwestern Polytechnical University
	\hspace{8pt}
	$^2$Inception Institute of Artificial Intelligence
	\\
	{\tt\small
    \{nnizhang.1995, junweihan2010, liunian228\}@gmail.com, ling.shao@ieee.org
    }
}

\maketitle
\footnotetext[1]{Corresponding author.} 
\ificcvfinal\thispagestyle{empty}\fi

\begin{abstract}
   Humans perform co-saliency detection by first summarizing the consensus knowledge in the whole group and then searching corresponding objects in each image. Previous methods usually lack robustness, scalability, or stability for the first process and simply fuse consensus features with image features for the second process.
   In this paper, we propose a novel consensus-aware dynamic convolution model to explicitly and effectively perform the ``summarize and search" process. To summarize consensus image features, we first summarize robust features for every single image using an effective pooling method and then aggregate cross-image consensus cues via the self-attention mechanism. By doing this, our model meets the scalability and stability requirements. Next, we generate dynamic kernels from consensus features
   to encode the summarized consensus knowledge. Two kinds of kernels are generated in a supplementary way to summarize fine-grained image-specific consensus object cues and the coarse group-wise common knowledge, respectively. Then, we can effectively perform object searching by employing dynamic convolution at multiple scales.
   Besides, a novel and effective data synthesis method is also proposed to train our network.
   Experimental results on four benchmark datasets verify the effectiveness of our proposed method. Our code and saliency maps are available at \url{https://github.com/nnizhang/CADC}. 
\end{abstract}

\section{Introduction}
\begin{figure}[!t]
  \graphicspath{{Figures/introduction/}}
  \centering
  \includegraphics[width=1\linewidth]{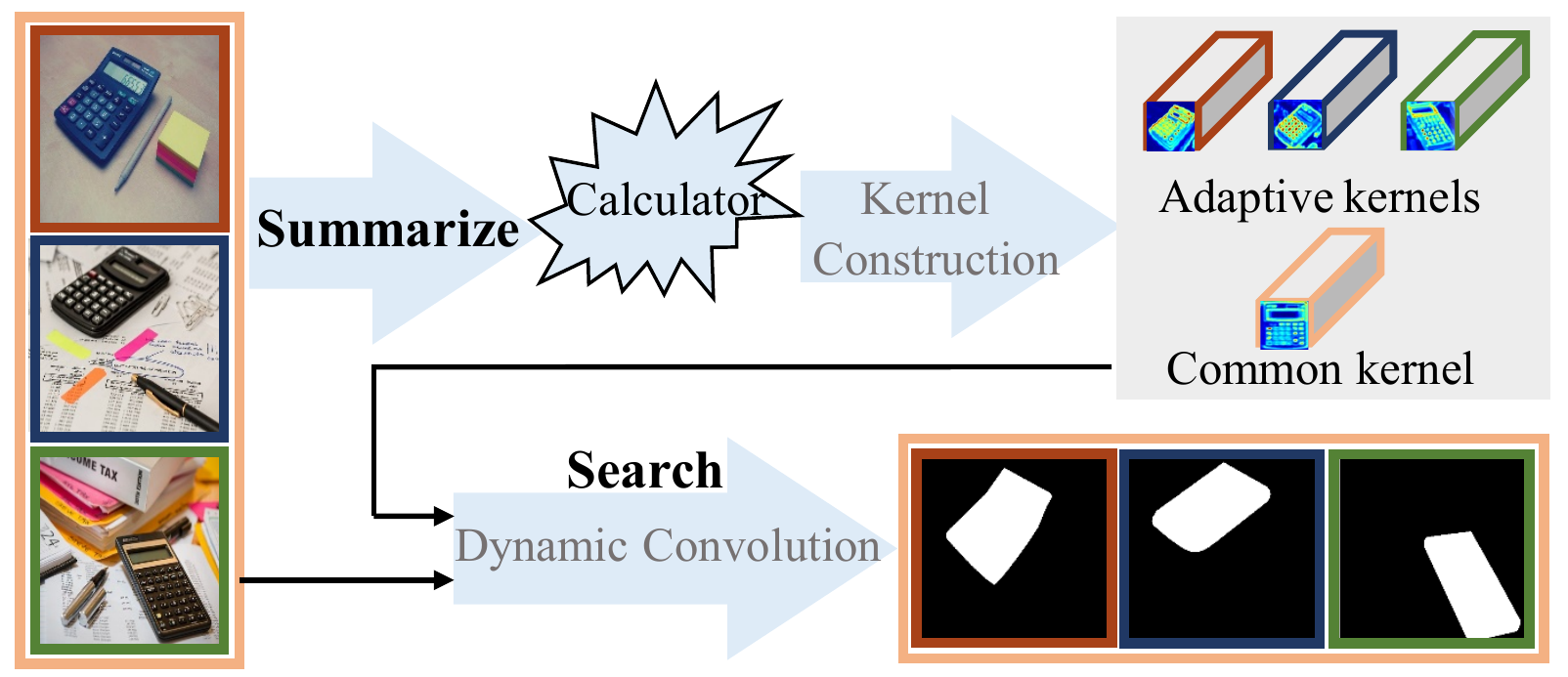}
  \caption{Main idea of our proposed method.}
  \label{figure1}
  \vspace{-0.3cm}
\end{figure}

Co-salient object detection (Co-SOD) mimics the human visual system to distinguish common and salient objects when viewing a group of relevant images.
Although various Co-SOD methods have been proposed, let us review this problem from the humans perspective.
Given a group of images, humans can not segment the co-salient object in each image directly. Instead, they need to first observe all images and summarize the consensus knowledge about what kind of objects this group is focusing on. Then, they look back at each image and search the corresponding objects. We call this process ``\emph{summarize and search}", which is illustrated in Figure~\ref{figure1}. A similar explanation can also be found in \cite{zhang2020gicd}. Therefore, we can model Co-SOD in such an intuitive way to summarize the consensus knowledge first and then search consensus objects in each image.

Previous models can also be explained from such a point. For consensus knowledge summarization,
early traditional methods employed graph models \cite{li2011co} or clustering methods \cite{fu2013cluster, yao2017revisiting} to learn the common patterns. However, their models lack end-to-end learning, thus limiting the model performance.
Some recent deep models \cite{wei2019deep, wei2017group, ren2020co} chose to concatenate and convolve all image features for summarizing the consensus knowledge. However, convolution can only aggregate the information at the same location among different images, while co-salient objects often show variations in scales and locations in different images. Hence, these models may easily fail in consensus summarization.
Using non-local dependencies \cite{wang2018non} to summarize the consensus cues is another choice \cite{gao2020co}.
However,
this method lacks scalability since it is computationally prohibitive for processing a large number of images.
Some other work \cite{li2019detecting} adopted recurrent networks to summarize the consensus cues step by step.
However, recurrent models define an input order for image sequences, thus lacking model stability since different input orders will lead to different results.

For consensus object searching,
many works \cite{wei2019deep, wei2017group, li2019detecting, wang2019robust, zha2020robust, ren2020co, zhang2020adaptive} directly fused the consensus feature with image-specific features via summation or concatenation operations.
\cite{zhang2019co} and \cite{deng2020re} fused co-attention maps with the image-specific information via element-wise multiplication.
Such simple methods conduct object searching by linear information fusion, which can not fully exploit the guidance of the summarized consensus knowledge.
Besides, \cite{zhang2020gicd}
computed channel-wise weight for each single image feature based on its similarity with the consensus representation,
which can be seen as an attribute-wise object searching method. We argue that direct spatial searching might be more 
accurate and easy to learn.

In this paper, we propose a novel consensus-aware dynamic convolution (CADC) model directly from the ``summarize and search" point of view.
The image features of the whole group are first summarized and then the consensus knowledge is encoded as dynamic kernels, which
capture the appearance traits of common objects.
Next, the searching step is performed by using the kernels to convolve the image features to obtain final results, as shown in Figure~\ref{figure1}.

However, adopting dynamic convolution for Co-SOD requires delicate model design. We propose to summarize the consensus knowledge via first summarizing the feature of every single image and then integrating cross-image consensus features. For the first step, we propose to use a multi-scale max-pooling module to achieve position and scale robust features. For the second step, we leverage the self-attention mechanism \cite{vaswani2017attention}. In this way, our model can meet the needs for scalability and stability.
For consensus-aware dynamic kernel generation, we propose to simultaneously construct image adaptive kernels and a common kernel. The former is generated for each image separately to capture fine-grained image-specific cues while the latter is generated for the whole group to summarize coarse group-wise common knowledge.
Theoretically, the latter can serve as a supplement and regularization for the former to avoid them focusing too much on the image-specific information. We also generate efficient large dynamic kernels to further consider spatial structures and enlarge the searching range.

Besides, considering the lack of training data in the Co-SOD field, we propose a novel and effective data synthesis method by fusing common objects with unrelated salient objects in two different ways to mimic the real-world scenarios. It can largely improve the Co-SOD performance and shows superiority when compared with previous methods.

Our main contributions can be summarized as follows.
\begin{compactitem}
\item From the ``summarize and search" perspective, we propose a novel CADC model for Co-SOD. Dynamic kernels are generated to summarize the consensus knowledge and object searching is performed using dynamic convolution.

\item We propose to combine multi-scale max-pooling and self-attention models to obtain consensus features with both model scalability and stability.

\item We construct two types of dynamic kernels in a supplementary way to capture image-specific cues and the group-wise common knowledge, respectively.

\item We develop a novel and more effective data synthesis method to mimic the challenging scenarios in the real world for Co-SOD models training.

\item Our CADC network achieves new state-of-the-art Co-SOD results.

\end{compactitem}

\section{Related Work}

\subsection{Co-Saliency Detection}
Early Co-SOD methods \cite{fu2013cluster, li2011co} often devote to design hand-crafted features based on different low-level image features.
Recent works have introduced deep learning techniques into Co-SOD and gained promising performance. One bunch of works \cite{zhang2016co,zhang2019co} combines deep learning features with other traditional methods. However, such separate learning schemes do not make full use of the advantage of CNNs in a data-driven manner.

In contrast, another bunch of works adopts end-to-end deep models to learn the common patterns of relevant images. \cite{wei2019deep, wei2017group, ren2020co} concatenated and convolved all image features to generate the consensus feature, which is sensitive to the variations of object locations and scales. In contrast, we propose a multi-scale max-pooling module to extract position and scale robust features. Besides, some works \cite{wei2019deep, wei2017group, li2019detecting, wang2019robust, zha2020robust, ren2020co, zhang2020adaptive} employed summation or concatenation operations to fuse the consensus features with single image features in a linear space. As a result, they can not explore more effective guidance from the consensus knowledge, hence performing unsatisfactorily for object searching. In contrast, we learn two types of consensus-aware dynamic kernels to perform diverse and supplementary consensus summarization and perform dynamic convolution for effective object searching.

\begin{figure*}[!t]
  \graphicspath{{Figures/Network/}}
  \centering
  \includegraphics[width=0.8\linewidth]{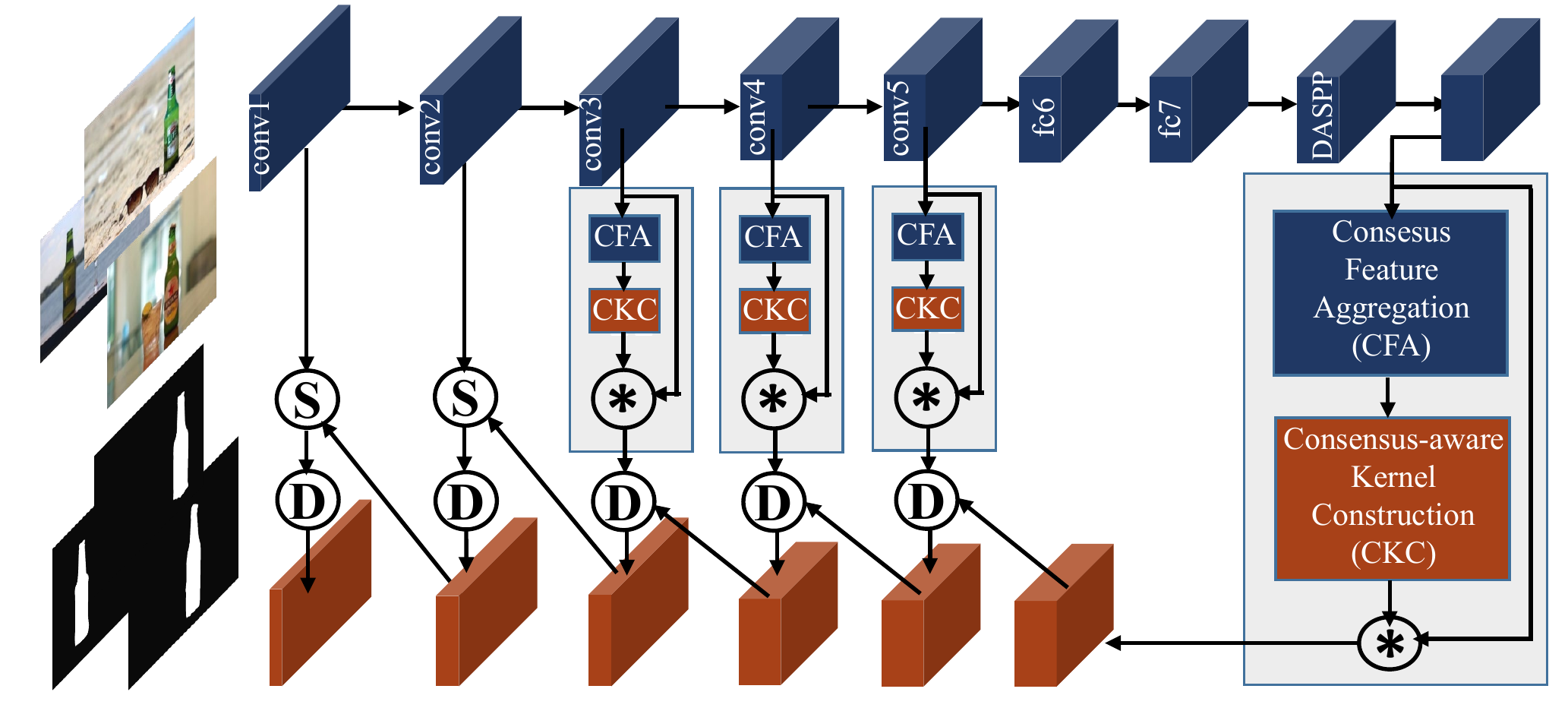}
  \caption{Framework of our CADC network. $\circledast$ and \textcircled{\small{D}} mean dynamic convolution and our decoder module, respectively. \textcircled{\small{S}} denotes the spatial attention.}
  \label{network_fig}
  \vspace{-0.3cm}
\end{figure*}

Some other existing models explore long-range dependencies to detect co-salient objects, such as \cite{gao2020co,li2019detecting}. However, \cite{gao2020co} only explored the interaction between a pair of images by the non-local network \cite{wang2018non}, which is fragile to obtain common features because similar extraneous objects may also appear in both images and cause distraction. Besides, this method also lacks scalability since it is computationally prohibitive to process a large number of images. \cite{li2019detecting} utilized recurrent networks to explore the interactions from all available images step by step. However, recurrent models have a sequential order issue and may cause model instability. In contrast, our adopted multi-scale max-pooling module can first decrease the feature dimensionality for each image, which further enables us to summarize the consensus knowledge from all images through the self-attention mechanism. Thus, we can effectively capture the global consensus with both model scalability and stability.

\subsection{Dynamic Convolution}
As a specified method of meta-learning, dynamic convolution uses predicted kernels to perform the convolution operation, which is different from traditional convolutions with fixed filters once trained. Xu \etal \cite{jia2016dynamic} proposed a dynamic filter network to learn custom parameters for different input samples. This idea is widely adopted to address the few-shot learning problem \cite{cai2018memory, gidaris2018dynamic}, where a learner is first trained on a large set of available training data of base categories and then utilized to generate dynamic weights for classifying novel categories.
Subsequently, some works \cite{qi2020pointins,tian2020conditional} introduced dynamic convolution into the instance segmentation task.
However, most of them only learned one dynamic kernel with the $1 \times 1$ size and did not consider learning large spatial kernels due to the involved large computational costs. Pang \etal \cite{HDFNetECCV2020} introduced large $3\times 3$ dynamic kernels with different dilation rates for RGB-D SOD. However, they generated a different kernel for every pixel in every image, which has significantly large computational costs. Different from previous methods, we design both group-specific and image-specific dynamic kernels to learn diverse and supplementary meta knowledge for Co-SOD. Our dynamic convolution is also computationally efficient by using the depthwise separable mechanism \cite{howard2017mobilenets}.

\section{Proposed Method}
Figure~\ref{network_fig} shows our overall pipeline for Co-SOD. We propose the CADC model for consensus summarization and object searching, where the former is performed by consensus feature aggregation and consensus-aware kernel construction. We embed this model into a U-shaped \cite{ronneberger2015unet} model and conduct hierarchical object searching in multiple feature levels.
At the same time, we propose a novel and effective data synthesis method to train the proposed network.

\begin{figure*}[!t]
  \graphicspath{{Figures/dynamic_kernel/}}
  \centering
  \includegraphics[width=1\linewidth]{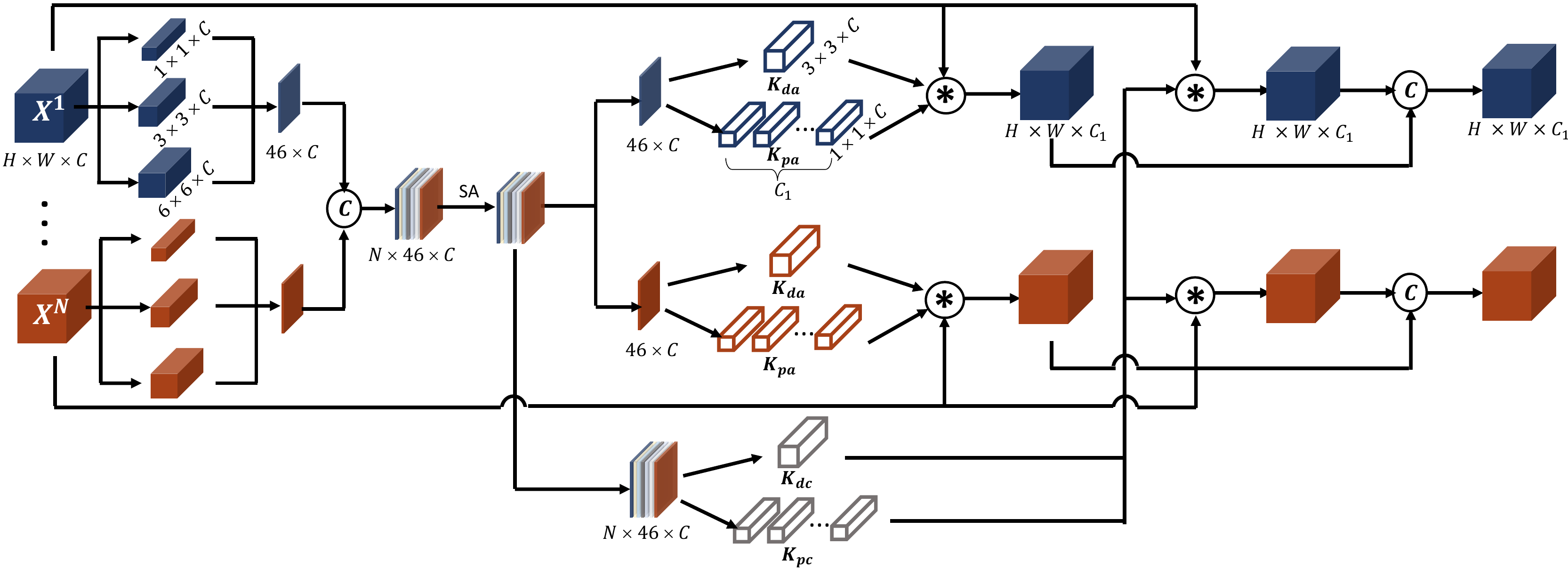}
  \caption{Pipeline of our proposed CADC for consensus summarization and object searching. We generate two types of kernels, \ie, adaptive kernel, and common kernel, for each image and the whole group, respectively. `SA' means the self-attention module. $\circledast$ and $\copyright$  mean the depthwise separable convolution and concatenation, respectively.}
  \label{kernel}
  \vspace{-0.3cm}
\end{figure*}

\subsection{Consensus Feature Aggregation}

Given a group of $N$ relevant images $\left\{I^n \right\}^N_{n=1}$, we first employ our encoder to extract their encoding feature maps $\bm{X}\in{\mathbb{R}^{N\times{H\times{W\times C}}}}$, where $H$, $W$, and $C$ represent its height, width, and channel number, respectively.
We follow \cite{liu2020S2MA} to slightly modify the original VGG-16 \cite{simonyan2014vgg} backbone and insert the modified DASPP module \cite{yang2018denseaspp} after it as our encoder.
Then, as shown in Figure~\ref{kernel}, we employ the adaptive max-pooling layer on $\bm{X}$ with three target scales and obtain the output features with spatial sizes of $1 \times 1$, $3 \times 3$ and $6\times 6$, respectively. Then, these output features are flattened and concatenated to generate a feature $\bm{F}\in{\mathbb{R}^{N\times{46\times C}}}$. For each image, the obtained feature summarizes the dominated object features at multiple scales, hence is robust to both position and scale variations of co-salient objects.

The multi-scale max-pooling module dramatically decreases the feature number of each image from $H\times W$ to 46, hence providing us the feasibility to summarize the global consensus from all images by self-attention.
Following \cite{vaswani2017attention}, we first apply linear transformations to project $\bm{F}$ into the query, key, and value spaces with $\frac{C}{2}$ channels. After that, an affinity matrix $\bm{A}\in{\mathbb{R}^{46N\times{46N}}}$ is calculated by matrix multiplication
between the query and the key matrices,
and it indicates the pairwise similarities among the $46N$ features of all images. Since a feature is usually more similar to other features in the same image than those in other images, we reset the self-similarity elements in $\bm{A}$ computed within each same image into a very small value to avoid intra-image similarities dominating the affinity matrix.

After that, we obtain an attention matrix via adopting normalization along the second dimension and then perform matrix multiplication with the value to generate an aggregated feature $\bm{Y}\in{\mathbb{R}^{46N\times{\frac{C}{2}}}}$.
Next, $\bm{Y}$ is re-projected to $C$ channels by a linear transformation and then reshaped to the shape $N\times{46\times C}$. Finally, it is added onto the original feature map $\bm{F}$ for providing a residual signal to generate the consensus feature $\bm{Z}\in{\mathbb{R}^{N\times{46\times C}}}$.

\subsection{Consensus-aware Kernel Construction}
Based on the consensus feature $\bm{Z}$, we generate two kinds of kernels for each image group to encode the summarized consensus knowledge. Since the co-occurring salient objects may have various appearances and scales in different images, we first construct adaptive kernels for each image to encode the fine-grained image-specific consensus object information. A common kernel is also generated for the whole group to capture the coarse group-wise common object knowledge. The latter can be regarded as a supplement and regularization for the former to avoid them paying too much attention to the image-specific information and ignoring the common information. To this end, generating these two kinds of kernels disentangles the learning of image-specific consensus object information and group-wise common object knowledge, thus better conforming to the nature of Co-SOD and facilitating potential relation exploration between them. Besides, doing so mimics the multi-branch architecture widely used in CNNs, which increases the transformation complexity and model capability.

\vspace{3mm}
\noindent\textbf{(1) Vanilla dynamic kernels with $1 \times 1$ size}
\vspace{2mm}

We first follow most traditional dynamic convolution methods \cite{qi2020pointins, tian2020conditional} to generate dynamic kernels with $ 1 \times 1$ size, which is straightforward and easy to implement.

\vspace{-3mm}
\paragraph{Adaptive kernels with ${1 \times 1}$ size.}
We utilize $\bm{Z}$ to generate adaptive kernels for different images. First, we flatten $\bm{Z}$ to $\mathbb{R}^{N\times 46C}$ and learn a feature attention $\bm{\alpha}\in{\mathbb{R}^{N\times46}}$ via:
\begin{equation} \label{1x1AdaKernelatt}
\bm{\alpha}=FC(ReLU(BN(FC(\bm{Z})))),
\end{equation}
where $\bm{\alpha}$ is further normalized by the softmax operation along the second dimension to select which one is the most discriminative among all the 46 features of each image. The intermediate FC layer has 1024 nodes.

Next, $\bm{Z}$ is weighted summed by $\bm{\alpha}$ along the second dimension to generate the attended feature $\bm{F_{a}}\in{\mathbb{R}^{N\times C}}$.

Finally, the ${1 \times 1}$ adaptive kernels are learned from $\bm{F_{a}}$ via:
\begin{equation} \label{1x1AdaKernelatt}
\bm{K_a}=FC(PReLU(BN(FC(\bm{F_{a}})))),
\end{equation}
where $\bm{K_{a}}\in{\mathbb{R}^{N\times C_1C}}$ and is further reshaped to the shape ${N \times {C_1\times{C\times{1\times 1}}}}$. Here $C_1$ is the desired output channel number of the dynamic convolution operation and the intermediate FC layer has $C$ nodes. We adopt the Parametric ReLU (PReLU) \cite{he2015delving} activation function for generating kernels since they usually have both positive and negative activation.

\vspace{-3mm}
\paragraph{Common Kernel with ${1 \times 1}$ size.}
We aim to employ an attention weight $\bm{W}\in{\mathbb{R}^{N \times 46}}$ to aggregate all image features in $\bm{Z}$ along the first two dimensions and generate a group-wise common feature $\bm{F_c}\in{\mathbb{R}^{C}}$.
As discussed in \cite{cao2019gcnet}, the computed self-attention of different queries all tend to highlight the same set of most discriminative key elements. Thus, we can find which features are the most discriminative from the self-attention matrix $softmax(\bm{A})$. Specifically, we can get the weight $\bm{W}$ by averaging this matrix along the first dimension. Then, $\bm{F_c}$ can be obtained by using $\bm{W}$ to weighted sum $\bm{Z}$ along the first two dimensions.

Finally, the common kernel can be learned via:
\begin{equation} \label{PoCoKernel}
\bm{K_c}=FC(PReLU(FC(\bm{F_c}))),
\end{equation}
where $\bm{K_{c}}\in{\mathbb{R}^{C_1C}}$ and is further reshaped to the shape $C_1 \times {C \times {1 \times 1}}$ as the group-wise kernel. The intermediate FC layer also has $C$ nodes.

\vspace{3mm}
\noindent\textbf{(2) Efficient large dynamic kernels}
\vspace{2mm}

The vanilla dynamic kernels with $1 \times 1$ size can only encode channel-wise consensus knowledge and ignore spatial cues. Besides, they can only enable object searching within the $1 \times 1$ range, resulting in limited searching capability. To introduce spatial cues for the consensus knowledge and also enlarge the searching range, we propose to generate large spatial dynamic kernels. However, it will incur large computation costs and large amounts of the FC parameters if directly use the same method as the vanilla dynamic kernels. For instance, if we want to generate dynamic kernels with a spatial size of $3 \times 3$, $\bm{K_a}$ and $\bm{K_c}$ will be 9 times larger, so do the parameters of the FC layers used to generate them.
This is also the reason that most dynamic convolution methods do not learn large spatial kernels. We overcome this issue by using the depthwise separable convolution \cite{howard2017mobilenets} operation. We construct both adaptive kernels and the common kernel with the $3 \times 3$ size in this form as follows.

\vspace{-3mm}
\paragraph{Adaptive Kernels with ${3 \times 3}$ size.}
We decompose $3 \times 3$ adaptive kernels $\bm{K_{la}}$ into depthwise adaptive kernels $\bm{K_{da}}\in{\mathbb{R}^{N \times{C\times{3\times 3}}}}$ and pointwise adaptive kernels $\bm{K_{pa}}\in{\mathbb{R}^{N \times{C_1\times{C\times{1\times 1}}}}}$. The latter can be constructed in the same way as $\bm{K_a}$.

To construct $\bm{K_{da}}$, We transform each of the 46-d features in $\bm{Z}$ to generate the $3 \times 3$ kernel for each channel and each image. We first permute $\bm{Z}$ to the shape $\mathbb{R}^{NC\times 46}$ and adopt FC layers as follows:
\begin{equation} \label{1x1AdaKernelatt}
\bm{K_{da}}=FC(PReLU(BN(FC(\bm{Z})))),
\end{equation}
where the intermediate FC layer has 46 nodes, $\bm{K_{da}}\in{\mathbb{R}^{NC \times 9}}$ and it is further reshaped to the shape $N \times{C \times {3 \times 3}}$.

\vspace{-3mm}
\paragraph{Common Kernel with ${3 \times 3}$ size.}
We also decompose the $3 \times 3$ common kernel into the depthwise common kernel $\bm{K_{dc}}\in{\mathbb{R}^{C\times{3\times 3}}}$ and the pointwise common kernel $\bm{K_{pc}}\in{\mathbb{R}^{C_1\times{C\times{1\times 1}}}}$. Note that the construction of $\bm{K_{pc}}$ is also the same as $\bm{K_c}$.

To construct $\bm{K_{dc}}$, we need to aggregate the information across $N$ images in the consensus feature $\bm{Z}$. To this end, we leverage an attention $\bm{\alpha_3}\in{\mathbb{R}^{N}}$ to aggregate the image features with its $N$ attention weights.

To learn $\bm{\alpha_3}$, we first flatten $\bm{Z}$ to the shape of $\mathbb{R}^{N\times 46C}$ and then generate two attentions $\bm{\alpha_1}\in{\mathbb{R}^{N \times C}}$ and $\bm{\alpha_2}\in{\mathbb{R}^{N \times 46}}$ via:
\begin{equation} \label{1x1AdaKernelatt}
\begin{split}
\bm{\alpha_1}=FC(ReLU(BN(FC(\bm{Z})))), \\
\bm{\alpha_2}=FC(ReLU(BN(FC(\bm{Z})))),
\end{split}
\end{equation}
where the intermediate FC layers for both $\bm{\alpha_1}$ and $\bm{\alpha_2}$ have 1024 nodes.
Then, $\bm{\alpha_1}$ and ${\bm{\alpha_2}}$ are further normalized by softmax along the second dimension. Next, we can obtain $\bm{\alpha_3}$ by successively using $\bm{\alpha_1}$ and $\bm{\alpha_2}$ to aggregate the information in $\bm{Z}$ along the ``$C$" and the ``46" dimensions.

Finally, we apply softmax on $\bm{\alpha_3}$ and employ it to weighted sum $\bm{Z}$ for eliminating the first dimension. As a result, we can obtain a feature $\bm{F_{dc}}\in{\mathbb{R}^{C \times 46}}$, which is utilized to generate $\bm{K_{dc}}$ via:
\begin{equation} \label{1x1AdaKernelatt}
\bm{K_{dc}}=FC(PReLU(BN(FC(\bm{F_{dc}})))),
\end{equation}
where the intermediate FC layer has 46 nodes, $\bm{K_{dc}}\in{\mathbb{R}^{C \times 9}}$ and is further reshaped to the shape $C\times {3 \times 3}$.

\subsection{Object Searching via Dynamic Convolution}
After obtaining consensus-aware dynamic kernels, we adopt dynamic convolution on the original feature maps $\left\{\bm{X^n} \right\}^N_{n=1}$ to perform explicit object searching. 
For vanilla $1 \times 1$ kernels, we directly use them to convolve each $\bm{X^n}$. 
For efficient large kernels, following depthwise separable convolution \cite{howard2017mobilenets}, we first use the depthwise kernels to perform $3 \times 3$ group convolution for each channel separately, and then adopt the pointwise kernels to perform regular $1 \times 1$ convolution.

For each image, we use both its adaptive kernel and the common kernel to perform dynamic convolution simultaneously and then fuse the two response maps to $C_1$ channels via concatenation and convolution, as shown in Figure~\ref{kernel}.

We use the proposed CADC to connect the encoder-decoder pairs in our U-shaped network at multiple levels, hence performing hierarchical object searching at different scales and can effectively improve the searching accuracy.
Specifically, we perform hierarchical object searching in the first four decoder modules.
In each decoder module, we first perform CADC on the encoder feature map. Then, we concatenate the searching response map with the previous decoder feature map and use two $3 \times 3$ Conv layers to fuse them. 
BN \cite{ioffe2015bn} layers and ReLU are also used right after the Conv layers.
For the last two decoder modules, we do not use CADC anymore. Instead, we simply use the previous decoder feature to generate a spatial attention map to filter the current encoder feature, as shown in Figure~\ref{network_fig}.

\begin{figure}[!t]
  \graphicspath{{Figures/synthesis_strategy/}}
  \centering
  \includegraphics[width=1\linewidth]{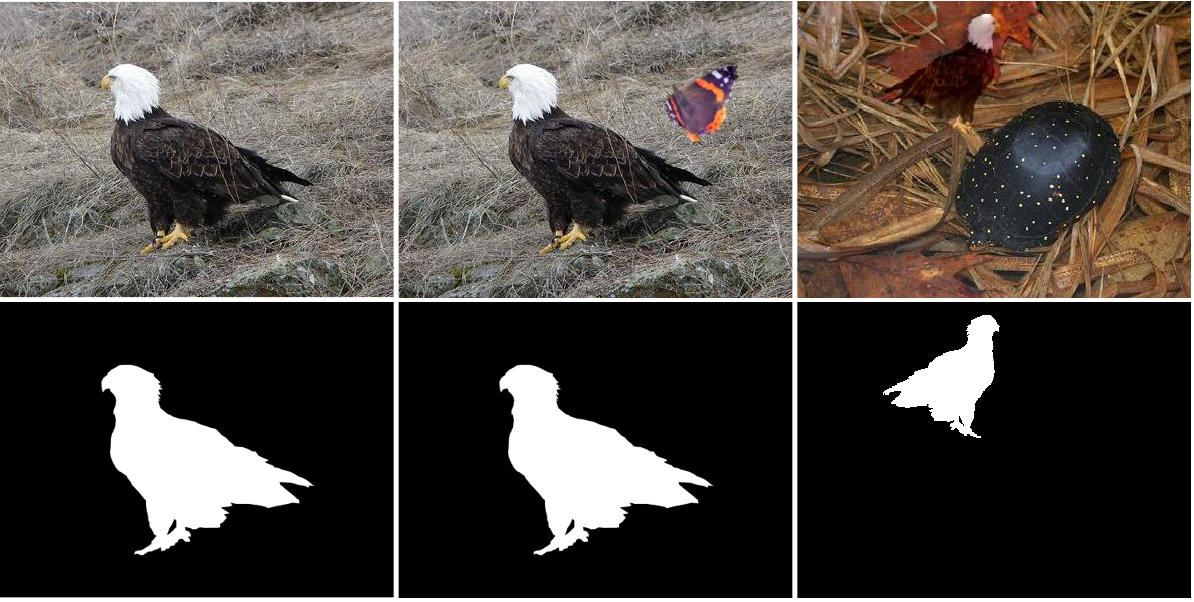}
  \caption{Examples of our proposed data synthesis method. The first and second columns show the original image and the normally synthesized image, and their corresponding ground truth. The last column shows the reversely synthesized image.}
  \label{synthesis}
  \vspace{-0.3cm}
\end{figure}

\subsection{Computational Costs Analysis}
In this section, we discuss the computational costs of our consensus feature aggregation and consensus-aware kernel construction.
In the former, our multi-scale max-pooling module dramatically decreases the feature number of each image from $H \times W$ to 46, hence making it possible to aggregate a group of images while the original self-attention incurs large computational costs.
For example, given N images, the computational complexity of using self-attention on the original feature maps and our pooled ones are $O((NWH)^{2})$ and $O((46N)^{2})$, respectively, where $46\ll WH$.
For the consensus-aware kernel construction, our model enlarges the searching range without dramatically increasing computational costs by introducing the depthwise-separable convolution.
It reduces the kernel size to construct from $C_1 \times {C \times {3 \times 3}}$ to $C \times {3 \times 3} + C_1 \times C$.

\section{New Data Synthesis Strategy}
Many previous models \cite{zhang2016co,han2017unified,li2018deep,li2019detecting} combined various datasets to train their Co-SOD models. We follow \cite{jin2020icnet} to use a subset of the COCO dataset \cite{lin2014microsoft} with 9213 images of 65 groups to train our model. However, this dataset highlights all objects that belong to the same category as ground truth, without discriminating salient and non-salient ones.
To this end, \cite{jin2020icnet} had to use an off-the-shelf SOD model \cite{zhao2019egnet} trained on DUTS \cite{wang2017learning} as a pre-computed saliency prior.
Hence, we also leverage DUTS \cite{wang2017learning} in our model training. 

To fit the DUTS dataset to the Co-SOD task, \cite{zhang2020gicd} divided its images into different groups based on the categories of salient objects, obtaining the DUTS class dataset, which contains 8250 images of 291 groups. However, each image in this dataset only contains target salient objects without distractions. To this end, \cite{zhang2020gicd} synthesized images for their model training by using a jigsaw strategy. This method splices each image of the target class with an image from other classes.
Although this strategy can simulate the distraction from extraneous salient objects in Co-SOD, it still has drawbacks that the splicing results are unnatural and objects will have large distortions when the synthesized images are resized to fixed shapes for network training.

Instead, we propose a copy and blend synthesis strategy based on poisson blending \cite{perez2003poisson}.
For each image of the target class, we randomly select an image from other classes and then copy and blend its salient object on the target image background as the distraction to generate the synthesized image.
However, for images synthesized in such a normal way, the target objects are usually more salient than the copied extraneous objects. As a result, the trained models easily downgrade to only learn to detect salient objects instead of co-salient objects. To tackle this issue, we also propose a reverse synthesis strategy to copy and blend target objects on the backgrounds of extraneous images using the same aforementioned synthesis method. Finally, we combine both normal and reverse strategies to train our model.

Compared with \cite{zhang2020gicd}, our proposed method can achieve more natural synthesis results and preserve reasonable shapes for objects, hence is more suitable for training Co-SOD models. Figure \ref{synthesis} shows some synthesized examples generated by our proposed method.

\section{Experiments}

\subsection{Datasets and Evaluation Metrics}

We evaluate our proposed method on four co-saliency benchmark datasets as follows. \textbf{MSRC} \cite{winn2005object} is collected for recognizing objects and we follow \cite{zhang2016detection, fu2013cluster} to select 233 images of 7 groups from MSRC for evaluation.
\textbf{CoSal2015} \cite{zhang2015co} and \textbf{CoSOD3k} \cite{fan2020taking} are two large-scale datasets containing 2015 images of 50 groups and 3316 images of 160 groups, respectively.
\textbf{CoCA} \cite{zhang2020gicd} is the latest and most challenging dataset which contains 1295 images of 80 groups. Different from other datasets, each image in CoCA contains at least one extraneous salient object, hence being more suitable for real-world applications and evaluating the performance of Co-SOD methods.

We adopt four widely-used evaluation metrics to compare our proposed method with state-of-the-art methods. Maximum F-measure (maxF) considers both precision and recall for co-saliency maps binarized by an optimal threshold. Structure-measure $S_m$ \cite{fan2017structure} considers object-aware and region-aware structural similarities. Enhanced-alignment measure $E_\xi$ \cite{Fan2018Enhanced} considers both global information and local details. Mean Absolute Error (MAE) computes the average absolute per-pixel difference between the predicted co-saliency maps and ground truth.

\subsection{Implementation Details}
In our data synthesis strategy, for each original image in DUTS class,
we generate three synthesized images using the normal strategy and another three using the reverse strategy.

We follow \cite{liu2018picanet} to conduct data augmentation and use $256\times 256$ as the training and testing size.
We employ the cross-entropy loss as the training loss and deploy deep supervision for each decoder module.
Stochastic gradient descent is used as our optimization algorithm.
We select at most 14 images from each group as each minibatch and set the total iteration step to 40,000. The initial learning rate is set to 0.01 and divided by 10 at the $20,000^{th}$ and the $30,000^{th}$ iterations, respectively.
Our code is implemented using Pytorch \cite{paszke2019pytorch}.

\begin{table}[t]
\centering
\renewcommand{\arraystretch}{1.0}
\renewcommand{\tabcolsep}{2.8mm}
\footnotesize
\caption{Quantitative results of different settings of our proposed model. ``VAK" and ``VCK" mean vanilla adaptive kernels and the vanilla common kernel, respectively. ``LAK" and ``LCK" represent large adaptive kernels and the large common kernel. ``ML" means adopting CADC at multiple decoder levels.}

\begin{tabular}{l|l|cccc}
\hline
\multicolumn{2}{l|}{\multirow{2}{*}{Settings}} & \multicolumn{4}{c}{CoCA} \\
\multicolumn{2}{l|}{}                      & \multicolumn{1}{l}{$S_m \uparrow$} & \multicolumn{1}{l}{maxF $\uparrow$} & \multicolumn{1}{l}{$E_\xi \uparrow$} & \multicolumn{1}{l}{MAE $\downarrow$}   \\ \hline

\multicolumn{2}{l|}{Baseline}  &0.633 &0.451 &0.707	&0.165  \\ \hline
\multicolumn{2}{l|}{+VAK}      &0.657 &0.495 &0.729	&0.153  \\
\multicolumn{2}{l|}{+VCK}      &0.655 &0.497 &0.711	&0.151   \\
\multicolumn{2}{l|}{+LAK}      &0.661 &0.508 &0.735	&0.146    \\
\multicolumn{2}{l|}{+LCK}      &0.659 &0.498 &0.722	&0.147\\
\multicolumn{2}{l|}{+LAK+LCK}  &0.665 &0.511 &0.731	&0.144  \\ \hline
\multicolumn{2}{l|}{+LAK+LCK+ML} &\textbf{0.681} &\textbf{0.548} &\textbf{0.744} &\textbf{0.132}\\ \hline
\end{tabular}
\label{kernel_ablation}
\end{table}

\begin{table}[]
\centering
\small
\caption{Quantitative results of using different training strategies.
}
\begin{tabular}{l|l|cccc}
\hline
\multicolumn{2}{l|}{\multirow{2}{*}{Train strategy}} & \multicolumn{4}{c}{CoCA} \\
\multicolumn{2}{l|}{}                      & \multicolumn{1}{l}{$S_m \uparrow$} & \multicolumn{1}{l}{maxF $\uparrow$} & \multicolumn{1}{l}{$E_\xi \uparrow$} & \multicolumn{1}{l}{MAE $\downarrow$}   \\ \hline

\multicolumn{2}{l|}{COCO-sub}    &0.628	&0.467 &0.707 &0.171   \\ \hline
\multicolumn{2}{l|}{+DUTS class \cite{zhang2020gicd}}  &0.645 &0.494 &0.720 &0.165 \\
\multicolumn{2}{l|}{+jigsaw strategy \cite{zhang2020gicd}}   &0.669	&0.537 &0.740 &0.149 \\ \hline
\multicolumn{2}{l|}{+normal strategy}                       &0.653 &0.504 &0.725 &0.157  \\
\multicolumn{2}{l|}{+reverse strategy}                      &0.653 &0.510 &0.735	&0.155\\
\multicolumn{2}{l|}{+bidirectional strategy}            &\textbf{0.681} &\textbf{0.548} &\textbf{0.744} &\textbf{0.132}\\ \hline
\end{tabular}
\label{strategy}
\end{table}

\subsection{Ablation Study}

We conduct ablation studies on the most challenging and the latest Co-SOD dataset CoCA \cite{zhang2020gicd}.

\paragraph{Effectiveness of CADC.}

\begin{figure}[t]
  \graphicspath{{Figures/ablation/}}
  \centering
  \includegraphics[width=1\linewidth]{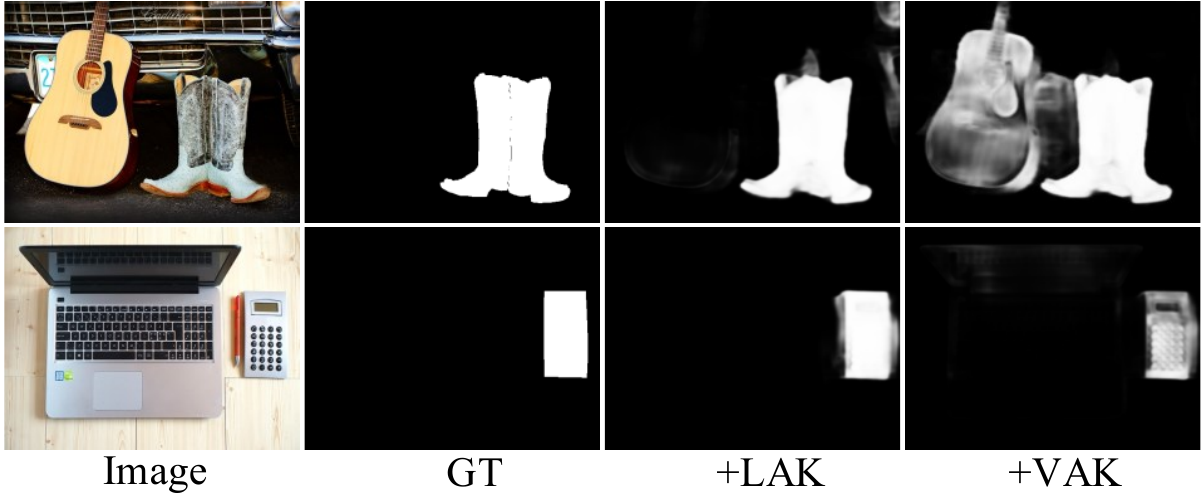}
  \caption{Visual comparison between ``+LAK" and ``+VAK".}
  \label{abl_large_range}
  \vspace{-0.3cm}
\end{figure}

\begin{figure}[t]
  \graphicspath{{Figures/ablation/}}
  \centering
  \includegraphics[width=1\linewidth]{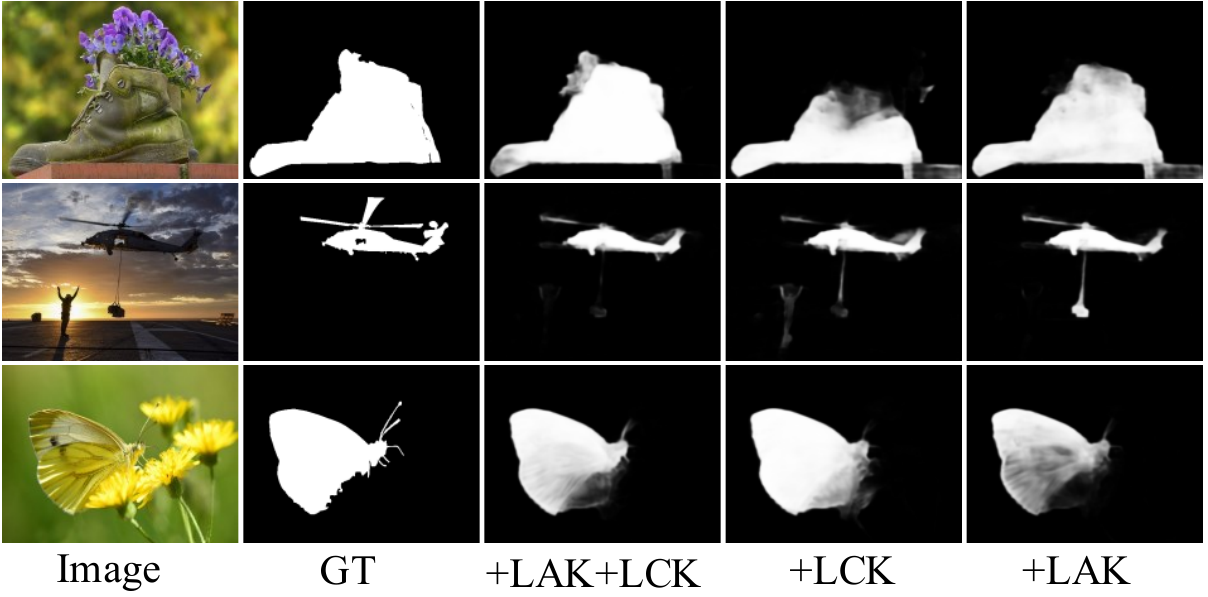}
  \caption{Visual comparison among ``+LAK", ``+LCK", and ``+LAK+LCK".}
  \label{ab_supplement}
  \vspace{-0.3cm}
\end{figure}

The first row in Table \ref{kernel_ablation} denotes our baseline model, \ie, employing UNet and DASPP with five simple decoders. This model degenerates to a pure SOD model without considering consensus information among images. Next, we separately use vanilla adaptive kernels (+VAK) and the vanilla common kernel (+VCK) in the baseline to incorporate consensus summarization. It can be seen that vanilla kernels obviously gain improvements when compared with the baseline. Furthermore, by adopting the efficient large dynamic kernels, \ie, large adaptive kernels (+LAK) and the large common kernel (+LCK), the model performance can be further improved when compared with using vanilla kernels. Figure~\ref{abl_large_range} also indicates that larger kernels can better search co-occurring objects while vanilla kernels can be easily interfered with by distraction objects or miss to completely segment the whole object. Combining these two kinds of kernels (+LAK+LCK) can bring more performance gains, indicating that consensus object searching can be better performed in a supplementary way. Figure~\ref{ab_supplement} also indicates adaptive kernels and common kernels can provide supplemental information.

Furthermore, we conduct hierarchical object searching on multiple levels (+LAK+LCK+ML), \ie, in the first four decoders. We can find that using dynamic convolution on multi-level feature maps can significantly bringing performance improvements.
Hence, we use this setting as our final CADC network.

\vspace{-3mm}

\begin{table*}[t]
  \centering
  \footnotesize
  \renewcommand{\arraystretch}{1.0}
  \renewcommand{\tabcolsep}{2.2mm}
 \caption{Quantitative comparison of our proposed model with other 11 SOTA Co-SOD methods on 4 benchmark datasets. \red{Red} and \blu{blue} denote the best and the second-best results, respectively. `-' indicates the code or result is not available.}
  \begin{tabular}{lr|ccccccccccc|c}
  \hline

    Dataset
    & Metric
    & CBCS  & DIM & CODW & MIL & IML  & SP-MIL & GONet & CSMG & GCAGC & GICD & ICNet & Ours \\
    &
    & \cite{fu2013cluster} &\cite{zhang2015cosaliency} & \cite{zhang2015co}   & \cite{zhang2015self} & \cite{ren2020co}& \cite{zhang2016co} & \cite{hsu2018unsupervised} & \cite{zhang2019co} & \cite{zhang2020adaptive} &\cite{zhang2020gicd} &\cite{jin2020icnet} &\\ \hline
  \multirow{4}{*}{CoCA}
    & $S_m\uparrow$    &0.526 &- &- &-	&-	&-	&-	&0.632	&-	&\blu{0.658}	&0.651	&\red{0.681}\\
    & maxF$\uparrow$  &0.315 &-	&-	&-	&-	&-	&-	&0.508	&-	&\blu{0.510}	&0.506	&\red{0.548}\\
    & $E_\xi\uparrow$  &0.638	&-	&-	&-	&-	&-	&-	&\blu{0.735}	&-	&0.712	&0.698	&\red{0.744}\\
  \cite{zhang2020gicd}& MAE$\downarrow$ &0.175	&-	&-	&-	&-	&-	&-	&\red{0.124}	&-	&\blu{0.125}	&0.148	&0.132\\
     \hline
  \multirow{4}{*}{CoSOD3k}
    & $S_m\uparrow$  &0.528	&0.559	&-	&-	&0.720	&-	&-	&0.711	&-	&0.778	&\blu{0.780}	&\red{0.801} \\
    & maxF$\uparrow$  &0.466 &0.495	&-	&-	&0.652	&-	&-	&0.709	&-	&\blu{0.744}	&\blu{0.744}	&\red{0.759}\\
    & $E_\xi\uparrow$ &0.637 &0.662	&-	&-	&0.773	&-	&-	&0.804	&-	&0.831	&\blu{0.832}	&\red{0.840}\\
  \cite{fan2020taking}& MAE$\downarrow$ &0.228 &0.327 &- &-	&0.164	&-	&-	&0.157	&-	&\red{0.089}	&0.097	&\blu{0.096}\\
     \hline
  \multirow{4}{*}{CoSal2015}
    & $S_m\uparrow$   &0.544 &0.592	&0.648 &0.673	&-	&-	&0.751	&0.774	&0.822	&0.842	&\blu{0.856} &\red{0.866} \\
    & maxF$\uparrow$  &0.532 &0.580	&0.667 &0.620	&-	&-	&0.740	&0.784	&0.843	&0.840	&\blu{0.855} &\red{0.862}\\
    & $E_\xi\uparrow$ &0.656 &0.695	&0.752 &0.720	&-	&-	&0.805	&0.842	&-	&0.885	&\blu{0.900} &\red{0.906} \\
    \cite{zhang2015co}& MAE$\downarrow$ &0.233 &0.312 &0.274 &0.210	&-	&-	&0.160	&0.130	&0.089	&0.071	&\red{0.058} &\blu{0.064}\\
     \hline
  \multirow{4}{*}{MSRC}
    & $S_m\uparrow$   &0.480 &0.657	&0.713 &0.720 &0.781 &0.769	&\blu{0.795} &0.722	&-	&0.665	&0.731	&\red{0.821} \\
    & maxF$\uparrow$  &0.630 &0.705 &0.784 &0.768 &0.840 &0.824 &0.846 &\blu{0.847} &- &0.692 &0.805 &\red{0.873}\\
    & $E_\xi\uparrow$  &0.676 &0.725 &0.820 &0.8 &0.856 &0.855 &\blu{0.863} &0.859 &-	&0.726 &0.822 &\red{0.895}\\
    \cite{winn2005object}& MAE$\downarrow$ &0.314 &0.309 &0.264 &0.216 &0.174 &0.218 &0.179 &0.190	&- &0.196 &\blu{0.160} &\red{0.115} \\
     \hline
  \end{tabular}
  \label{SOTA}
\end{table*}

\begin{figure*}[t]
  \graphicspath{{Figures/qualitative/}}
  \centering
  \includegraphics[width=1\linewidth]{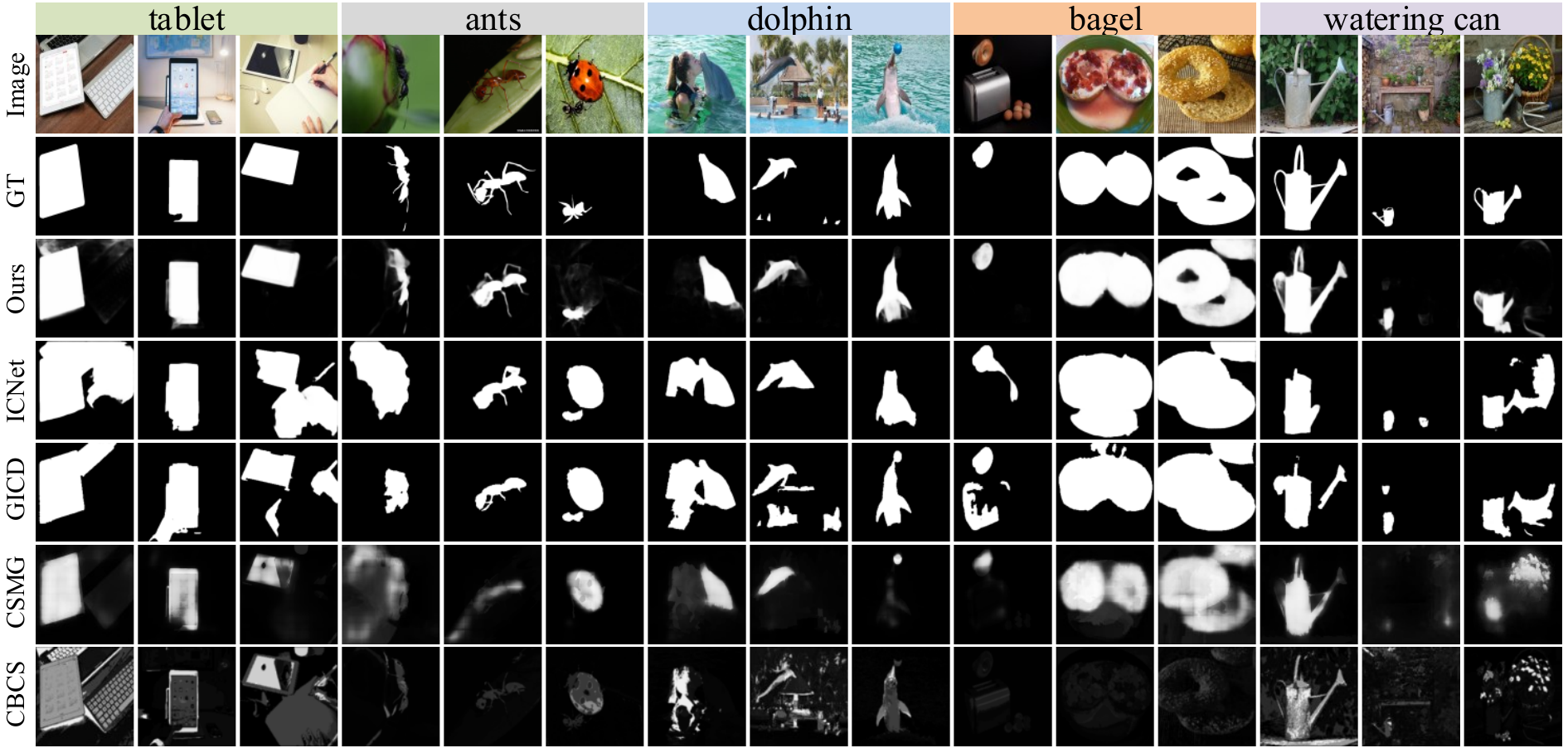}
  \caption{Qualitative comparisons of our proposed model with other state-of-the-art methods.}
  \label{qualitative}
  \vspace{-0.3cm}
\end{figure*}

\vspace{-3mm}
\paragraph{Effectiveness of our data synthesis strategy.}
Table \ref{strategy} shows the comparison results of training our model on different data.
We first train our model on the COCO subset, \ie, COCO-sub. Then, we respectively add the original DUTS class dataset, the synthesized images using the jigsaw strategy \cite{zhang2020gicd}, synthesizing only using our normal synthesis strategy, synthesizing only using our reverse synthesis strategy, and using our bidirectional synthesis strategy (normal and reverse).
The results show that adding original DUTS class images can bring performance gains compared to only using the COCO-sub dataset, indicating the supplementary of saliency attributes is necessary. 
Furthermore, only using our normal or reverse strategy can obtain slightly better results than using the original DUTS class data. However, using both of them can lead to large performance gains and outperform the jigsaw strategy. Hence, our normal and reverse synthesis strategies provide complementary cues to each other and both of them are indispensable for effective model training.

\subsection{Comparison with State-of-the-Art Methods}
We compare our proposed model with other 11 state-of-the-art methods, \ie, CBCS \cite{fu2013cluster}, DIM \cite{zhang2015cosaliency}, CODW \cite{zhang2015co}, MIL \cite{zhang2015self}, IML \cite{ren2020co}, SP-MIL \cite{zhang2016co}, GONet \cite{hsu2018unsupervised}, CSMG \cite{zhang2019co}, GCAGC \cite{zhang2020adaptive}, GICD \cite{zhang2020gicd} , and ICNet \cite{jin2020icnet}.

We illustrate the quantitative comparison results in table \ref{SOTA}. Generally, our model achieves the best performance on all four datasets.
On the most challenging dataset CoCA, our model brings 3.8\% improvement in terms of maxF compared with the second-best method.
We also show qualitative comparison results in Figure \ref{qualitative}. It can be seen that our model can better search and segment the co-occurring salient objects in many challenging scenes while other methods often are disturbed by other extraneous salient objects.
Specifically, for the ants class, our model can accurately search the targets which are similar to the background while other methods either lost the targets or be interfered with by other salient objects.

\section{Conclusion}
In this paper, we propose a consensus-aware dynamic convolution model to explicitly perform the ``summarize and search" process for co-saliency detection. Two types of efficient large dynamic kernels are constructed in a supplementary way to capture image-specific consensus object cues and the group-wise common knowledge, respectively.
We hierarchically search the co-salient objects by performing the dynamic convolution operation at multiple levels.
We also present a new data synthesis method to effectively mimic the distraction of extraneous objects in the real world. Extensive experimental results demonstrate the effectiveness of our proposed method.

\vspace{-4mm}
\paragraph{Acknowledgments:}
This work was supported in part by the National Key R\&D Program of China under Grant 2018YFB1402600, the National Science Foundation of China under Grant 62027813, 62036005, U20B2065, U20B2068.

{\small
\bibliographystyle{ieee_fullname}
\bibliography{egbib}

\begin{thebibliography}{10}\itemsep=-1pt

\bibitem{cai2018memory}
Qi Cai, Yingwei Pan, Ting Yao, Chenggang Yan, and Tao Mei.
\newblock Memory matching networks for one-shot image recognition.
\newblock In {\em CVPR}, pages 4080--4088, 2018.

\bibitem{cao2019gcnet}
Yue Cao, Jiarui Xu, Stephen Lin, Fangyun Wei, and Han Hu.
\newblock Gcnet: Non-local networks meet squeeze-excitation networks and
  beyond.
\newblock In {\em ICCV Workshops}, 2019.

\bibitem{fan2017structure}
Deng-Ping Fan, Ming-Ming Cheng, Yun Liu, Tao Li, and Ali Borji.
\newblock Structure-measure: A new way to evaluate foreground maps.
\newblock In {\em ICCV}, pages 4548--4557, 2017.

\bibitem{Fan2018Enhanced}
Deng-Ping Fan, Cheng Gong, Yang Cao, Bo Ren, Ming-Ming Cheng, and Ali Borji.
\newblock {Enhanced-alignment Measure for Binary Foreground Map Evaluation}.
\newblock In {\em IJCAI}, pages 698--704, 2018.

\bibitem{deng2020re}
Deng-Ping Fan, Tengpeng Li, Zheng Lin, Ge-Peng Ji, Dingwen Zhang, Ming-Ming
  Cheng, Huazhu Fu, and Jianbing Shen.
\newblock Re-thinking co-salient object detection.
\newblock {\em TPAMI}, 2021.

\bibitem{fan2020taking}
Deng-Ping Fan, Zheng Lin, Ge-Peng Ji, Dingwen Zhang, Huazhu Fu, and Ming-Ming
  Cheng.
\newblock Taking a deeper look at co-salient object detection.
\newblock In {\em CVPR}, pages 2919--2929, 2020.

\bibitem{fu2013cluster}
Huazhu Fu, Xiaochun Cao, and Zhuowen Tu.
\newblock Cluster-based co-saliency detection.
\newblock {\em TIP}, 22(10):3766--3778, 2013.

\bibitem{gao2020co}
Guangshuai Gao, Wenting Zhao, Qingjie Liu, and Yunhong Wang.
\newblock Co-saliency detection with co-attention fully convolutional network.
\newblock {\em IEEE Transactions on Circuits and Systems for Video Technology},
  31(3):877--889, 2020.

\bibitem{gidaris2018dynamic}
Spyros Gidaris and Nikos Komodakis.
\newblock Dynamic few-shot visual learning without forgetting.
\newblock In {\em CVPR}, pages 4367--4375, 2018.

\bibitem{han2017unified}
Junwei Han, Gong Cheng, Zhenpeng Li, and Dingwen Zhang.
\newblock A unified metric learning-based framework for co-saliency detection.
\newblock {\em IEEE Transactions on Circuits and Systems for Video Technology},
  28(10):2473--2483, 2017.

\bibitem{he2015delving}
Kaiming He, Xiangyu Zhang, Shaoqing Ren, and Jian Sun.
\newblock Delving deep into rectifiers: Surpassing human-level performance on
  imagenet classification.
\newblock In {\em ICCV}, pages 1026--1034, 2015.

\bibitem{howard2017mobilenets}
Andrew~G Howard, Menglong Zhu, Bo Chen, Dmitry Kalenichenko, Weijun Wang,
  Tobias Weyand, Marco Andreetto, and Hartwig Adam.
\newblock Mobilenets: Efficient convolutional neural networks for mobile vision
  applications.
\newblock {\em arXiv preprint arXiv:1704.04861}, 2017.

\bibitem{hsu2018unsupervised}
Kuang-Jui Hsu, Chung-Chi Tsai, Yen-Yu Lin, Xiaoning Qian, and Yung-Yu Chuang.
\newblock Unsupervised cnn-based co-saliency detection with graphical
  optimization.
\newblock In {\em ECCV}, pages 485--501, 2018.

\bibitem{ioffe2015bn}
Sergey Ioffe and Christian Szegedy.
\newblock Batch normalization: Accelerating deep network training by reducing
  internal covariate shift.
\newblock In {\em ICML}, pages 448--456, 2015.

\bibitem{jia2016dynamic}
Xu Jia, Bert De~Brabandere, Tinne Tuytelaars, and Luc~V Gool.
\newblock Dynamic filter networks.
\newblock In {\em NIPS}, pages 667--675, 2016.

\bibitem{jin2020icnet}
Wen-Da Jin, Jun Xu, Ming-Ming Cheng, Yi Zhang, and Wei Guo.
\newblock Icnet: Intra-saliency correlation network for co-saliency detection.
\newblock {\em NIPS}, 2020.

\bibitem{li2019detecting}
Bo Li, Zhengxing Sun, Lv Tang, Yunhan Sun, and Jinlong Shi.
\newblock Detecting robust co-saliency with recurrent co-attention neural
  network.
\newblock In {\em IJCAI}, pages 818--825, 2019.

\bibitem{li2011co}
Hongliang Li and King~Ngi Ngan.
\newblock A co-saliency model of image pairs.
\newblock {\em TIP}, 20(12):3365--3375, 2011.

\bibitem{li2018deep}
Min Li, Shizhong Dong, Kun Zhang, Zhifan Gao, Xi Wu, Heye Zhang, Guang Yang,
  and Shuo Li.
\newblock Deep learning intra-image and inter-images features for co-saliency
  detection.
\newblock In {\em BMVC}, volume 291, 2018.

\bibitem{lin2014microsoft}
Tsung-Yi Lin, Michael Maire, Serge Belongie, James Hays, Pietro Perona, Deva
  Ramanan, Piotr Doll{\'a}r, and C~Lawrence Zitnick.
\newblock Microsoft coco: Common objects in context.
\newblock In {\em ECCV}, pages 740--755, 2014.

\bibitem{liu2018picanet}
Nian Liu, Junwei Han, and Ming-Hsuan Yang.
\newblock Picanet: Learning pixel-wise contextual attention for saliency
  detection.
\newblock In {\em CVPR}, pages 3089--3098, 2018.

\bibitem{liu2020S2MA}
Nian Liu, Ni Zhang, and Junwei Han.
\newblock Learning selective self-mutual attention for rgb-d saliency
  detection.
\newblock In {\em CVPR}, pages 13756--13765, 2020.

\bibitem{HDFNetECCV2020}
Youwei Pang, Lihe Zhang, Xiaoqi Zhao, and Huchuan Lu.
\newblock Hierarchical dynamic filtering network for rgb-d salient object
  detection.
\newblock In {\em ECCV}, pages 235--252, 2020.

\bibitem{paszke2019pytorch}
Adam Paszke, Sam Gross, Francisco Massa, Adam Lerer, James Bradbury, Gregory
  Chanan, Trevor Killeen, Zeming Lin, Natalia Gimelshein, Luca Antiga, et~al.
\newblock Pytorch: An imperative style, high-performance deep learning library.
\newblock {\em NIPS}, 32:8026--8037, 2019.

\bibitem{perez2003poisson}
Patrick P{\'e}rez, Michel Gangnet, and Andrew Blake.
\newblock Poisson image editing.
\newblock In {\em ACM SIGGRAPH 2003 Papers}, pages 313--318. 2003.

\bibitem{qi2020pointins}
Lu Qi, Yi Wang, Yukang Chen, Ying-Cong Chen, Xiangyu Zhang, Jian Sun, and Jiaya
  Jia.
\newblock Pointins: Point-based instance segmentation.
\newblock {\em TPAMI}, 2021.

\bibitem{ren2020co}
Jingru Ren, Zhi Liu, Xiaofei Zhou, Cong Bai, and Guangling Sun.
\newblock Co-saliency detection via integration of multi-layer convolutional
  features and inter-image propagation.
\newblock {\em Neurocomputing}, 371:137--146, 2020.

\bibitem{ronneberger2015unet}
Olaf Ronneberger, Philipp Fischer, and Thomas Brox.
\newblock U-net: Convolutional networks for biomedical image segmentation.
\newblock In {\em MICCAI}, pages 234--241, 2015.

\bibitem{simonyan2014vgg}
Karen Simonyan and Andrew Zisserman.
\newblock Very deep convolutional networks for large-scale image recognition.
\newblock In {\em ICLR}, 2015.

\bibitem{tian2020conditional}
Zhi Tian, Chunhua Shen, and Hao Chen.
\newblock Conditional convolutions for instance segmentation.
\newblock In {\em ECCV}, pages 282--298, 2020.

\bibitem{vaswani2017attention}
Ashish Vaswani, Noam Shazeer, Niki Parmar, Jakob Uszkoreit, Llion Jones,
  Aidan~N Gomez, {\L}ukasz Kaiser, and Illia Polosukhin.
\newblock Attention is all you need.
\newblock In {\em NIPS}, pages 5998--6008, 2017.

\bibitem{wang2019robust}
Chong Wang, Zheng-Jun Zha, Dong Liu, and Hongtao Xie.
\newblock Robust deep co-saliency detection with group semantic.
\newblock In {\em AAAI}, volume~33, pages 8917--8924, 2019.

\bibitem{wang2017learning}
Lijun Wang, Huchuan Lu, Yifan Wang, Mengyang Feng, Dong Wang, Baocai Yin, and
  Xiang Ruan.
\newblock Learning to detect salient objects with image-level supervision.
\newblock In {\em CVPR}, pages 136--145, 2017.

\bibitem{wang2018non}
Xiaolong Wang, Ross Girshick, Abhinav Gupta, and Kaiming He.
\newblock Non-local neural networks.
\newblock In {\em CVPR}, pages 7794--7803, 2018.

\bibitem{wei2017group}
Lina Wei, Shanshan Zhao, Omar El~Farouk Bourahla, Xi Li, and Fei Wu.
\newblock Group-wise deep co-saliency detection.
\newblock In {\em IJCAI}, pages 3041--3047, 2017.

\bibitem{wei2019deep}
Lina Wei, Shanshan Zhao, Omar El~Farouk Bourahla, Xi Li, Fei Wu, and Yueting
  Zhuang.
\newblock Deep group-wise fully convolutional network for co-saliency detection
  with graph propagation.
\newblock {\em TIP}, 28(10):5052--5063, 2019.

\bibitem{winn2005object}
John Winn, Antonio Criminisi, and Thomas Minka.
\newblock Object categorization by learned universal visual dictionary.
\newblock In {\em ICCV}, volume~2, pages 1800--1807, 2005.

\bibitem{yang2018denseaspp}
Maoke Yang, Kun Yu, Chi Zhang, Zhiwei Li, and Kuiyuan Yang.
\newblock Denseaspp for semantic segmentation in street scenes.
\newblock In {\em CVPR}, pages 3684--3692, 2018.

\bibitem{yao2017revisiting}
Xiwen Yao, Junwei Han, Dingwen Zhang, and Feiping Nie.
\newblock Revisiting co-saliency detection: A novel approach based on two-stage
  multi-view spectral rotation co-clustering.
\newblock {\em TIP}, 26(7):3196--3209, 2017.

\bibitem{zha2020robust}
Zheng-Jun Zha, Chong Wang, Dong Liu, Hongtao Xie, and Yongdong Zhang.
\newblock Robust deep co-saliency detection with group semantic and pyramid
  attention.
\newblock {\em TNNLS}, 31(7):2398--2408, 2020.

\bibitem{zhang2015cosaliency}
Dingwen Zhang, Junwei Han, Jungong Han, and Ling Shao.
\newblock Cosaliency detection based on intrasaliency prior transfer and deep
  intersaliency mining.
\newblock {\em TNNLS}, 27(6):1163--1176, 2015.

\bibitem{zhang2015co}
Dingwen Zhang, Junwei Han, Chao Li, and Jingdong Wang.
\newblock Co-saliency detection via looking deep and wide.
\newblock In {\em CVPR}, pages 2994--3002, 2015.

\bibitem{zhang2016detection}
Dingwen Zhang, Junwei Han, Chao Li, Jingdong Wang, and Xuelong Li.
\newblock Detection of co-salient objects by looking deep and wide.
\newblock {\em IJCV}, 120(2):215--232, 2016.

\bibitem{zhang2016co}
Dingwen Zhang, Deyu Meng, and Junwei Han.
\newblock Co-saliency detection via a self-paced multiple-instance learning
  framework.
\newblock {\em TPAMI}, 39(5):865--878, 2016.

\bibitem{zhang2015self}
Dingwen Zhang, Deyu Meng, Chao Li, Lu Jiang, Qian Zhao, and Junwei Han.
\newblock A self-paced multiple-instance learning framework for co-saliency
  detection.
\newblock In {\em ICCV}, pages 594--602, 2015.

\bibitem{zhang2019co}
Kaihua Zhang, Tengpeng Li, Bo Liu, and Qingshan Liu.
\newblock Co-saliency detection via mask-guided fully convolutional networks
  with multi-scale label smoothing.
\newblock In {\em CVPR}, pages 3095--3104, 2019.

\bibitem{zhang2020adaptive}
Kaihua Zhang, Tengpeng Li, Shiwen Shen, Bo Liu, Jin Chen, and Qingshan Liu.
\newblock Adaptive graph convolutional network with attention graph clustering
  for co-saliency detection.
\newblock In {\em CVPR}, pages 9050--9059, 2020.

\bibitem{zhang2020gicd}
Zhao Zhang, Wenda Jin, Jun Xu, and Ming-Ming Cheng.
\newblock Gradient-induced co-saliency detection.
\newblock In {\em ECCV}, pages 455--472, 2020.

\bibitem{zhao2019egnet}
Jia-Xing Zhao, Jiang-Jiang Liu, Deng-Ping Fan, Yang Cao, Jufeng Yang, and
  Ming-Ming Cheng.
\newblock Egnet: Edge guidance network for salient object detection.
\newblock In {\em ICCV}, pages 8779--8788, 2019.

\end{thebibliography}
}

\end{document}